\documentclass[runningheads]{llncs}
\usepackage{etex}

\usepackage{times}
\usepackage{epsfig}
\usepackage{graphicx}
\usepackage{tabularx}
\usepackage{booktabs}
\usepackage{array}
\usepackage{color}
\usepackage[width=122mm,left=12mm,paperwidth=146mm,height=193mm,top=12mm,paperheight=217mm]{geometry}

\usepackage{amsmath}
\usepackage{bbm}
\usepackage{bm}
\usepackage{amssymb}
\usepackage{mathabx}
\usepackage[pagebackref=true,breaklinks=true,letterpaper=true,colorlinks,bookmarks=false]{hyperref}
\usepackage{lipsum}
\usepackage{multirow}
\usepackage{url}
\usepackage{lettrine}
\usepackage{cite}
\usepackage[ruled,vlined,linesnumbered]{algorithm2e}
\usepackage{pgfplots}
\usepackage{subcaption}
\usepackage{pgfplotstable}
\usepackage{pgf,tikz}
\usetikzlibrary{positioning}

\hypersetup{%
  bookmarksnumbered,
  pdfstartview={FitH},
  colorlinks,
  citecolor=black,
  filecolor=black,
  linkcolor=black,
  urlcolor=black,
  breaklinks=true,
}

\begin{document}
\pagestyle{headings}
\mainmatter

\title{LOH and Behold: Web-scale visual search, recommendation and clustering using Locally Optimized Hashing}
\titlerunning{LOH and behold!}

\authorrunning{Kalantidis \emph{et al.}}
\author{Yannis Kalantidis$^\dagger$, Lyndon Kennedy$^\ddagger$\thanks{Work done while author was at Yahoo Labs.}, Huy Nguyen$^\dagger$, \\ Clayton Mellina$^\dagger$ and David A. Shamma$^{\S\star}$}

\institute{$^\dagger$Computer Vision and Machine Learning Group, Flickr, Yahoo\\
$^\ddagger$Futurewei Technologies Inc. \\
$^\S$CWI: Centrum Wiskunde \& Informatica, Amsterdam \\
\vspace{1pc}
\email{\{ykal,huyng,clayton\}@yahoo-inc.com, \\ lyndonk@acm.org, aymans@acm.org}
}

\maketitle

\newcommand{\nospace}{\vspace{-4mm}}
\newcommand{\head}[1]{{\smallskip\noindent\bf{#1}}}
\newcommand{\alert}[1]{{\color{red}{#1}}}


\newcommand{\one}{\mathbbm{1}}
\newcommand{\bbF}{\mathbb{F}}
\newcommand{\bbH}{\mathbb{H}}
\newcommand{\expect}{\mathbb{E}}
\newcommand{\project}{\mathbb{P}}
\newcommand{\real}{\mathbb{R}}
\newcommand{\tran}{^{\mathrm{T}}}
\newcommand{\diff}{\mathrm{d}}
\newcommand{\prob}{\mathrm{Pr}}
\newcommand{\binomial}{\mathrm{Bi}}
\newcommand{\weibull}{\mathrm{Wb}}
\newcommand{\normal}{\mathcal{N}}
\newcommand{\wishart}{\mathcal{W}}
\newcommand{\zcol}{\mathbf{0}}
\newcommand{\zrow}{\zcol\tran}
\newcommand{\otherwise}{\mathrm{otherwise}}

\newcommand{\cA}{\mathcal{A}}
\newcommand{\cB}{\mathcal{B}}
\newcommand{\cC}{\mathcal{C}}
\newcommand{\cD}{\mathcal{D}}
\newcommand{\cE}{\mathcal{E}}
\newcommand{\cF}{\mathcal{F}}
\newcommand{\cG}{\mathcal{G}}
\newcommand{\cH}{\mathcal{H}}
\newcommand{\cI}{\mathcal{I}}
\newcommand{\cJ}{\mathcal{J}}
\newcommand{\cK}{\mathcal{K}}
\newcommand{\cL}{\mathcal{L}}
\newcommand{\cM}{\mathcal{M}}
\newcommand{\cN}{\mathcal{N}}
\newcommand{\cO}{\mathcal{O}}
\newcommand{\cP}{\mathcal{P}}
\newcommand{\cQ}{\mathcal{Q}}
\newcommand{\cR}{\mathcal{R}}
\newcommand{\cS}{\mathcal{S}}
\newcommand{\cT}{\mathcal{T}}
\newcommand{\cU}{\mathcal{U}}
\newcommand{\cV}{\mathcal{V}}
\newcommand{\cW}{\mathcal{W}}
\newcommand{\cX}{\mathcal{X}}
\newcommand{\cY}{\mathcal{Y}}
\newcommand{\cZ}{\mathcal{Z}}

\newcommand{\va}{\mathbf{a}}
\newcommand{\vb}{\mathbf{b}}
\newcommand{\vc}{\mathbf{c}}
\newcommand{\vd}{\mathbf{d}}
\newcommand{\ve}{\mathbf{e}}
\newcommand{\vf}{\mathbf{f}}
\newcommand{\vg}{\mathbf{g}}
\newcommand{\vh}{\mathbf{h}}
\newcommand{\vi}{\mathbf{i}}
\newcommand{\vj}{\mathbf{j}}
\newcommand{\vk}{\mathbf{k}}
\newcommand{\vl}{\mathbf{l}}
\newcommand{\vm}{\mathbf{m}}
\newcommand{\vn}{\mathbf{n}}
\newcommand{\vo}{\mathbf{o}}
\newcommand{\vp}{\mathbf{p}}
\newcommand{\vq}{\mathbf{q}}
\newcommand{\vr}{\mathbf{r}}
\newcommand{\vt}{\mathbf{t}}
\newcommand{\vu}{\mathbf{u}}
\newcommand{\vv}{\mathbf{v}}
\newcommand{\vw}{\mathbf{w}}
\newcommand{\vx}{\mathbf{x}}
\newcommand{\vy}{\mathbf{y}}
\newcommand{\vz}{\mathbf{z}}

\newcommand{\vA}{\mathbf{A}}
\newcommand{\vB}{\mathbf{B}}
\newcommand{\vC}{\mathbf{C}}
\newcommand{\vD}{\mathbf{D}}
\newcommand{\vF}{\mathbf{F}}
\newcommand{\vG}{\mathbf{G}}
\newcommand{\vI}{\mathbf{I}}
\newcommand{\vM}{\mathbf{M}}
\newcommand{\vS}{\mathbf{S}}
\newcommand{\vW}{\mathbf{W}}
\newcommand{\vX}{\mathbf{X}}
\newcommand{\vzero}{\mathbf{0}}

\newcommand{\rLambda}{\mathrm{\Lambda}}
\newcommand{\rSigma}{\mathrm{\Sigma}}

\newcommand{\vmu}{\bm{\mu}}
\newcommand{\vpi}{\bm{\pi}}
\newcommand{\vLambda}{\bm{\rLambda}}
\newcommand{\vSigma}{\bm{\rSigma}}
\newcommand{\indicator}{\mathbbm{1}}

%
\newcommand{\eg}{\emph{e.g. }} \newcommand{\Eg}{\emph{E.g}}
\newcommand{\ie}{\emph{i.e. }} \newcommand{\Ie}{\emph{I.e}}
\newcommand{\etc}{\emph{etc}} 
\newcommand{\etal}{\emph{et al. }}


\newcommand{\plot}[2]{\cffolderinput{#1}{#2}}
\newcommand{\data}[1]{\cfcurrentfolder #1.txt}
\newcommand{\tabread}[2]{\pgfplotstableread{\data{#1}}#2}

\newcommand{\leg}[1]{\addlegendentry{#1}}
\newcommand{\legtext}[1]{\addlegendimage{empty legend} \addlegendentry{#1}}

\tikzset{every mark/.append style={solid}}
\pgfplotsset{ grid=both, width=\columnwidth, try min ticks=5,
	 every axis x label/.style={at={(ticklabel cs:0.5)},anchor=north},
	 every axis y label/.style={at={(ticklabel cs:0.5)},rotate=90,anchor=south},
	 every axis/.append style={font=\scriptsize,thick,mark=*,mark size=1.3, smooth,tension=0.18},
	 legend cell align=left, legend style={fill opacity=0.8}
}

\pgfplotsset{
    dash/.style={mark=o,dashed,opacity=0.6},
    mmark/.style={mark=*},
}
\newcommand{\kilo}[1]{\thisrow{#1}/1000}


\begin{abstract}

We propose a novel hashing-based matching scheme, called Locally Optimized Hashing (LOH), based on a state-of-the-art quantization algorithm that can be used for efficient, large-scale search, recommendation, clustering, and deduplication. We show that matching with LOH only requires set intersections and summations to compute and so is easily implemented in generic distributed computing systems. We further show application of LOH to: a) large-scale search tasks where performance is on par with other state-of-the-art hashing approaches; b) large-scale recommendation where queries consisting of thousands of images can be used to generate accurate recommendations from collections of hundreds of millions of images; and c) efficient clustering with a graph-based algorithm that can be scaled to massive collections in a distributed environment or can be used for deduplication for small collections, like search results, performing better than traditional hashing approaches while only requiring a few milliseconds to run. In this paper we experiment on datasets of up to 100 million images, but in practice our system 
can scale to larger collections and can be used for other types of data that have a vector representation in a Euclidean space.

\end{abstract}


\section{Introduction}
\label{sec:intro}

\noindent
The rapid rise in the amount of visual multimedia created, shared, and consumed requires the development of better large-scale methods for querying and mining large data collections. Similarly, with increased volume of data comes a greater variety of use cases, requiring simple and repurposeable pipelines that can flexibly adapt to growing data and changing requirements.

Recent advances in computer vision have shown a great deal of progress in analyzing the content of very large image collections, pushing the state-of-the-art for classification~\cite{RDS+14}, detection~\cite{DABP14, GDDM14} and visual similarity search~\cite{KNA+07, ZZS+09, AKTS10, KTA+11, ToJA15}. 
Critically, deep Convolutional Neural Networks (CNNs)~\cite{KrSH12} have allowed processing pipelines to become much simpler by reducing complex engineered systems to simpler systems learned end-to-end and by providing powerful, generic visual representations that can be used for a variety of downstream visual tasks. Recently it has been shown that such deep features can be used to reduce visual search to nearest neighbor search in the deep feature space~\cite{BSCL14}. Complimentary work has recently produced efficient algorithms for approximate nearest neighbor search that can scale to billions of vectors~\cite{KaAv14, JTDA11, GHKS14}. 

In this paper, we present a novel matching signature, called \emph{Locally Optimized Hashing} (\emph{LOH}). LOH extends LOPQ~\cite{KaAv14}, a state-of-the-art nearest neighbor search algorithm, by treating the quantization codes of LOPQ as outputs of hashing functions. When applied to deep features, our algorithm provides a very flexible solution to a variety of related large-scale search and data mining tasks, including fast visual search, recommendation, clustering, and deduplication. Moreover, unlike~\cite{KaAv14, JTDA11, GHKS14}, our system does not necessarily require specialized resources (\ie dedicated cluster nodes and indexes for visual search) and is easily implemented in generic distributed computing environments.

Our approach sacrifices precision for speed and generality as compared to more exact quantization approaches, but it enables applications that wouldn't be computationally feasible with more exact approaches. LOH can trivially cope with large multi-image query sets. In practice, our approach allows datasets of \textit{hundreds of millions} of images to be efficiently searched with query sets of \emph{thousands of images}. We are in fact able to query with multiple large query sets, \eg from Flickr groups, simultaneously and get visual recommendations for all the sets in parallel. We are also able to cluster web-scale datasets with MapReduce by simply thresholding LOH matches and running a connected components algorithm. The same approach can be used for deduplication of, \eg search results.

Our contributions can be summarized as follows:
\begin{enumerate}
\item We propose Locally Optimized Hashing (LOH), a novel hashing-based matching method that competes favorably with the state-of-the-art hashing methods for search and allows approximate ordering of results. 
\item We extend LOH to multiple image queries and provide a simple and scalable algorithm that can provide visual recommendations in batch for query sets of thousands of images.
\item We show that this same representation can be used to efficiently deduplicate image search results and cluster collections of hundreds of millions of images. 
\end{enumerate}

Although in this paper we experiment on datasets of up to 100 million images (\ie using the YFCC100M dataset~\cite{TES+16}, the largest publicly available image dataset), in practice our system is suited to web-scale multimedia search applications with billions of images. In fact, on a Hadoop cluster with 1000 nodes, our approach can find and rank similar images\emph{ for millions of users from a search set of hundreds of millions of images in a runtime on the order of one hour.} The method can be adapted to other data types that have vector representations in Euclidean space.

\section{Related Work}
\label{sec:related}


\noindent
Large scale nearest neighbor search was traditionally based on
hashing methods~\cite{DIIM04, PaJA10} because they offer low memory footprints for index codes and fast search in Hamming space~\cite{NoPF12}. However, even recent hashing approaches~\cite{JHL+13, WaKC10, SSLS15} suffer in terms of performance compared to quantization-based approaches~\cite{JeDS11,NoPF12} for the same amount of memory.  On the other hand, quantization-based approaches traditionally performed worse in terms of search times, and it was only recently with the use of novel indexing methods~\cite{BaLe12} that quantization-based search was able to achieve search times of a few
milliseconds in databases of billions of vectors~\cite{GHKS14, KaAv14, NoFl13}.


A benefit of quantization approaches is that, unlike classic hashing methods, they provide a ranking for the retrieved points. Recently, approaches for binary code re-ranking have been proposed in~\cite{WSYY14, ZZT+13}; both papers propose a secondary, computationally heavier re-ranking step that, although is performed on only the retrieved points, makes search slower than state-of-the-art quantization-based approaches.  In the approach presented here, we try to keep the best of both worlds by producing an approximate ordering of retrieved points without re-ranking. We argue that for use cases involving multiple queries, this approximation can be tolerated since many ranked lists are aggregated in this case. 

A similar approach to ours, \ie an approach that aims to produce multipurpose, \textit{polysemous} codes~\cite{DoJP16} is presented at the current ECCV conference. After training a product quantizer, the authors then propose to optimize the so-called index assignment of the centroids to binary codes, such that distances between similar centroids are small in the Hamming space.

For multi-image queries, there are two broad categories based on the semantic concepts that the query image set represents. The first is query sets that share the same semantic concept or even the same specific object (\ie a particular building in Oxford)~\cite{ArZi12b, FeTu13, ZhHS14, ToJA15}. The second category is multi-image queries with \emph{multiple} semantics. This category has been recently studied~\cite{HsCA14} and the authors propose a Pareto-depth approach on top Efficient Manifold Ranking~\cite{XBC+11} for such queries. Their approach is however not scalable to very large databases and they limit query sets to just be image pairs.  

The current work uses visual features from a CNN trained for classification, thus similarities in our visual space capture broader category-level semantics. We focus on the first category of multi-image queries, \ie multiple-image query sets with a single semantic concept, and provide a simple and scalable approach which we apply to Flickr group set expansion. However, it is straightforward to tackle the second category with our approach by introducing a first step of (visual or multi-modal) clustering on the query set with multiple semantics before proceeding with the LOH-based set expansion.



\section{Locally Optimized Hashing}
\label{sec:loh}


\subsection{Background}
\label{subsec:background}

\subsubsection{Product quantization.}
A \emph{quantizer} $q$ maps a $d$-dimensional vector $\vx \in \real^d$ to vector $q(\vx) \in \cC$, where $\cC$ is a finite subset of $\real^d$, of cardinality $k$. Each vector $\vc \in \cC$ is called a \emph{centroid}, and $\cC$ a \emph{codebook}. Assuming that dimension $d$ is a multiple of $m$, we may write any vector $\vx \in \real^d$ as a concatenation $(\vx^1, \dots, \vx^m)$ of $m$ sub-vectors, each of dimension $d/m$. If $\cC^1, \dots, \cC^m$ are $m$ sub-codebooks of $k$ sub-centroids in subspace $\real^{d/m}$, a \emph{product quantizer}~\cite{JeDS11} constrains $\cC$ to be a Cartesian product
\begin{equation}
	\cC = \cC^1 \times \dots \times \cC^m,
\label{eq:cartesian}
\end{equation}
making it  a codebook of $k^m$ centroids of the form $\vc = (\vc^1, \dots, \vc^m)$ with each sub-centroid $\vc^j \in \cC^j$ for $j \in \cM = \{ 1, \dots, m \}$. An optimal product quantizer $q$ should minimize distortion $E = \sum_{\vx \in \cX} \| \vx - q(\vx) \|^2.$ as a function of $\cC$, subject to $\cC$ being of the form~(\ref{eq:cartesian})~\cite{GHKS14}. This is typically done with a variant of $k$-means.

When codebook $\cC$ is expressed as a Cartesian product, for each vector $\vx \in \real^d$, the nearest centroid in $\cC$ is 
\begin{equation}
q(\vx) = (q^1(\vx^1), \dots, q^m(\vx^m)),
\label{eq:subquantizers}
\end{equation}
where $q^j(\vx^j)$ is the nearest sub-centroid of sub-vector $\vx^j$ in $\cC^j$, for $j \in \cM$~\cite{GHKS14}. Hence finding an optimal product quantizer $q$ in $d$ dimensions amounts to solving $m$ optimal sub-quantizer problems $q^j, j \in \cM$, each in $d/m$ dimensions.

Given a new \emph{query} vector $\vy$, the (squared) Euclidean distance to every point $\vx \in \cX$ may be approximated by
\begin{equation}
	\delta^{SDC}(\vy, \vx) = \sum_{j=1}^m \| q^j(\vy^j) - q^j(\vx^j) \|^2,
\label{eq:sdc}
\end{equation}
or 
\begin{equation}
	\delta^{ADC}(\vy, \vx) = \sum_{j=1}^m \| \vy^j - q^j(\vx^j) \|^2,
\label{eq:adc}
\end{equation}

where $q^j(\vx^j) \in \cC^j = \{ \vc_1^j, \dots, \vc_k^j \}$ for $j \in \cM$. The superscripts $SDC$ and $ADC$ correspond to the symmetric and asymmetric distance computations of~\cite{JeDS11}, respectively. In the latter case the query vector is not quantized using the product quantizer, distances $\| \vy^j - \vc_i^j \|^2$ are computed and stored for $i \in \cK$ and $j \in \cM$ prior to search, so~(\ref{eq:adc}) amounts to only $O(m)$ operations. 
Sacrificing distortion for speed, in this approach we are mostly exploring the symmetric approximation~(\ref{eq:sdc}), where the query is also in quantized form. In this case, sub-quantizer distances $\| \vc_l^j - \vc_i^j \|^2$ can be pre-computed and stored for all $i,l \in \cK$ and $j \in \cM$ and again only $O(m)$ operations are needed for distance computations.


\subsubsection{Locally Optimized Product Quantization.}
In their recent paper~\cite{KaAv14}, the authors further extend product quantization by optimizing multiple product quantizers \emph{locally}, after some initial, coarse quantization of the space. Similar to the IVFADC version of~\cite{JeDS11} or multi-index~\cite{BaLe12}, they adopt a two-stage quantization scheme,  where local optimization follows independently inside each cluster of a coarse quantizer $Q$, learnt on the residual vectors with respect to the cluster's centroid.
They learn an \emph{optimized} product quantizer~\cite{GHKS14} per cluster, jointly optimizing the subspace decomposition together with the sub-quantizers. Constraint~(\ref{eq:cartesian}) of the codebook is relaxed to
\begin{equation}
	\cC = \{ R\hat{\vc} : \hat{\vc} \in \cC^1 \times \dots \times \cC^m, R\tran R = I\},
\label{eq:rotate}
\end{equation}
where the $d \times d$ matrix $R$ is orthogonal and allows for arbitrary rotation and permutation of vector components. 

Given the coarse quantizer $Q$ and the associated codebook $\cE = \{ \ve_1, \dots, \ve_K \}$ of $K$ \textit{clusters}, for $i \in \cK = \{ 1, \dots, K \}$ we may define the set of residuals of all data points $\vx \in \cX_i$ quantized to cluster $i$ as $\cZ_i = \{ \vx - \ve_i : \vx \in \cX_i, \ Q(\vx) = \ve_i \}$.  Given a set $\cZ \in \{ \cZ_1, \dots, \cZ_K \}$, the problem of locally optimizing both space decomposition and sub-quantizers can be expressed as minimizing distortion as a function of orthogonal matrix $R \in \real^{d \times d}$ and sub-codebooks $\cC^1, \dots, \cC^m \subset \real^{d / m}$ per cell,
\begin{equation}
	\begin{array}{rl}
	  \mathrm{minimize}    & {\displaystyle \sum_{\vz \in \cZ} \min_{\hat{\vc} \in \hat{\cC}} \| \vz - R\hat{\vc} \|^2} \\
	  \mathrm{subject\ to} & \hat{\cC} = \cC^1 \times \dots \times \cC^m \\
	                       & R\tran R = I,
	\end{array}
\label{eq:opt}
\end{equation}
where $|\cC^j| = k$ for $j \in \cM = \{ 1, \dots, m \}$. Assuming a $d$-dimensional, zero-mean normal distribution $\cN(\vzero, \Sigma)$ of residual data $\cZ$, we can efficiently solve the problem by first aligning the data with PCA and then using the \emph{eigenvalue allocation}~\cite{GHKS14} algorithm to assign dimensions to subspaces.

To achieve the state-of-the-art results on a billion-scale dataset, the inverted multi-index~\cite{BaLe12} is used with local optimization. 
In this setting, the original space is split into two 
subspaces first and then LOPQ follows within each one of the two subspaces separately, on the residual vectors. 

Each data point now gets assigned in \textit{two} clusters, one in each subspace. The intersection of two clusters gives a multi-index \textit{cell} in the product space.
However, as the space overhead to locally optimize per cell is prohibitive, in ~\cite{KaAv14} the authors separately optimize \textit{per cluster} in each of the two subspaces. They refer to this type of local optimization as \textit{product optimization} and the complete algorithm as \emph{Multi-LOPQ}, which is the approach we also adopt for training. 
As was shown in~\cite{KaAv14}, using local rotations together with a single set of global sub-quantizers gave only a small drop in performance. We therefore choose to have a global set of sub-quantizers $q(\vx)$ defined as in (\ref{eq:subquantizers}) and trained on the projected residual vectors $\vx$ from (\ref{eq:residual}).
We will refer to the two quantization stages as \textit{coarse} and \textit{fine} quantization, respectively.

\subsection{Locally Optimized Hashing}
\label{subsec:loh}

\noindent
The symmetric distance computation of (\ref{eq:sdc}) yields an approximation of the true distance in the quantized space. In pursuit of an even more scalable, fast and distributed-computing friendly approach, we propose to treat the sub-quantizer centroid indices as \emph{hash codes}. This allows us to approximate the true distance via \emph{collisions} without any explicit numeric computations apart from a final summation. As we demonstrate below, the proposed formulation further generalizes to querying with multiple vector queries, and, in fact, can perform many such queries in parallel very efficiently.

Lets begin by assuming a single coarse quantizer $Q$. The LOPQ model contains local rotations $R_c$ and global sub-quantizers $q_c^j, j \in \cM$ and $c \in \cK$ that operate on the projected residuals. Let $\vx, \vy \in \cZ_c$ be such residual vectors with respect to the same centroid of cluster $c$ of the coarse quantizer. The symmetric distance computation of~(\ref{eq:sdc}) is now given by

\begin{equation}
	\delta^{SDC}_c(\vy, \vx) = \| q_c(\vy) - q_c(\vx) \|^2 = \sum_{j=1}^m \| q_c^j(\vy^j) - q_c^j(\vx^j) \|^2,
\label{eq:sdc2}
\end{equation}
where $q_c^j$ is the $j$-th sub-quantizer for cluster $c$, with $c \in \cK$, $j \in \cM$ and $\vx, \vy \in \cZ_c$.  A residual vector $\vx$  is mapped via $q_c$ to the corresponding sub-centroid indices $i_c(\vx) = (i_c(\vx)^1, \dots, i_c(\vx)^m)$. 
We may treat the indices as values of a set of hash functions $h_c = (h_c^1, \dots,h_c^m)$, \ie a  mapping $h_c:  \real^d \rightarrow \mathbb{Z}^m$ such that $h_c(\vx) = i_c(\vx)$. 
We can then estimate the \emph{similarity} between residual vectors $\vy, \vx$ using the function:
\begin{equation}
	\sigma_h(\vy, \vx) = \sum_{j=1}^m  \indicator [ h_c(\vy)^j = h_c(\vx)^j ],
\label{eq:simh}
\end{equation}
where $\indicator [a = b]$ equals to $1$, iff $a = b$ and $0$ otherwise. Our hash functions are defined locally, \ie on residual vectors for a specific cluster, and therefore we call the proposed matching scheme \emph{Locally Optimized Hashing} or \emph{LOH}.

The LOH approach can be extended to work on top of a multi-LOPQ~\cite{KaAv14} model. Instead of having a single coarse quantizer,  we learn two subspace quantizers $Q^1, Q^2$ of $K$ centroids, with associated codebooks $\cE^j = \{ \ve_1^j, \dots, \ve_K^j \}$ for $j = 1, 2$ in a product quantization fashion.  

Each data point $\tilde{\vx} = (\tilde{\vx}^1, \tilde{\vx}^2) \in \cX$ is quantized using the two coarse quantizers into the tuple 
\begin{equation}
(c_1,c_2) = (\arg \min_i \| \tilde{\vx}^1 - \ve_i^1 \|, \arg \min_i \| \tilde{\vx}^2 - \ve_i^2 \|),
\end{equation}
with $c_j \in [1,K]$ referring to the indices of the nearest clusters for the two subspaces $j=1,2$. We will refer to the tuple $c(\tilde{\vx}) = (c_1,c_2)$ as the \emph{coarse codes} of a data point. Following LOPQ, the residual vector $\vx = (\vx^1, \vx^2)$ of point $\tilde{\vx}$ is equal to:
\begin{equation}
\vx = (R_{c_1}^1(\tilde{\vx}^1 - \ve_{c_1}^1), R_{c_2}^2(\tilde{\vx}^2 - \ve_{c_2}^2) ),
\label{eq:residual}
\end{equation}
where $R_i^j$ correspond to the local rotation of cluster $i$ in subspace $j$. We can split the concatenated vector $\vx$ into $m$ subvectors $\vx = (\vx^1, \ldots, \vx^m)$ and use the global sub-quantizers for encoding into $m$ codes. Therefore, given sub-quantizer $q(\vx)$ with codebooks $\vc = (\vc^1, \dots, \vc^m)$ and each sub-centroid $\vc^j \in \cC^j =  \{ \vc_1^j, \dots, \vc_k^j \}$ for $j \in \cM = \{ 1, \dots, m \}$, we can compute
\begin{equation}
f_j = \arg \min_i \| \vx^j - \vc_i^j \|, 
\end{equation}
for all $j \in \cM$ and get the sub-quantizer indices for the $m$ subspaces in set $f(\vx) = (f_1, \ldots, f_m)$. We will refer to this set as the \textit{fine codes} of a data point. An overview of the encoding process is shown in Algorithm~\ref{alg:encode}.

Now, given residual vectors $\vx, \vy$ with respect to the same set of coarse codes, and their sets of fine codes $f(\vx), f(\vy)$, the similarity function of (\ref{eq:simh}) can be expressed as
\begin{eqnarray}
\sigma_h(\vy, \vx) = \sum_{i=1}^{m}  \indicator [ f_i(\vy) = f_i(\vx) ] 
\label{eq:sim_mi}
\end{eqnarray}
\ie the sum of similarities for the $m$ subspaces. We should note that since fine codes are calculated on residuals, \ie given a coarse centroid, they are comparable only for points that share at least one of the coarse codes. If two points, for example share only the first coarse code of the two, only the first half of the fine codes are comparable. 


\begin{algorithm}[t]
\footnotesize{
\DontPrintSemicolon
\SetFuncSty{textsc}
\SetDataSty{emph}
\newcommand{\commentsty}[1]{{\color{green!50!black}#1}}
\SetCommentSty{commentsty}
\SetKwInOut{Input}{input}
\SetKwInOut{Output}{output}
\Input{data point $\tilde{\vx} \in \cX$, number of subspaces $m$, coarse quantizer codebooks $\cE^j$, local rotations $R_i^j$ where $ i=1,\ldots,K$ and $j=1,2$ and sub-quantizer codebooks  $\vc = (\vc^1, \dots, \vc^m)$.
}
\Output{sets of coarse and fine codes}
\BlankLine
\tcp{calculate coarse codes}
$ (c_1,c_2) =  (\arg \min_i \| \tilde{\vx}^1 - \ve_i^1 \|, \arg \min_i \| \tilde{\vx}^2 - \ve_i^2 \| )$
\BlankLine

\tcp{calculate locally projected residuals}
$ \vx = (R_{c_1}^1(\tilde{\vx}^1 - \ve_{c_1}^1), R_{c_2}^2(\tilde{\vx}^2 - \ve_{c_2}^2) )$ 
\BlankLine

\tcp{split residual to $m$ subvectors}
$\vx = (\vx^1, \ldots, \vx^m)$
\BlankLine

\tcp{calculate fine codes}
$(f_1, \ldots, f_m) = ( \arg \min_i \| \vx^1 - \vc_i^1 \|, \ldots,  \arg \min_i \| \vx^m - \vc_i^m \|)$ 
\BlankLine
\caption{Data encoding for the multi-index case}
\label{alg:encode}
}
\end{algorithm}

\subsection{Ranking with LOH}
\label{subsec:search}

\noindent
After we encode all database points using the approach summarized in Algorithm~\ref{alg:encode}, each one will be assigned to the \textit{cell} of the multi-index that corresponds to the pair of coarse codes of each data point. For indexing, an inverted list $\cL_c$ is kept for each cell $c = c_{ij} = (c_i, c_j)$ of the multi-index, giving $K^2$ inverted lists in total. 


For search one visits the inverted lists of multiple cells. The query vector is therefore not projected to just the closest coarse cluster for each of the coarse quantizers, but to multiple, and cells are visited in a sequence dictated by the multi-sequence algorithm~\cite{BaLe12}. As we are only counting collisions of fine codes within a coarse cluster, we need a way of incorporating the distance of the query to the centroid for each cell visited to further rank points \textit{across} cells. The multi-sequence algorithm provides us with an (approximate) distance $d_c$ for each cell which we use to extend the similarity function presented in the previous section. We introduce a weight $w_c$ for each cell $c$ visited that encodes the similarity of the query's residual to that cell's centroid, and modify the similarity function to be:

\begin{eqnarray}
	\sigma_w(\vy, \vx) &=& w_c + \sigma_h(\vy, \vx) = w_c + \sum_{i=1}^{m}  \indicator [ f_i(\vy) = f_i(\vx) ] 
\label{eq:sim}
\end{eqnarray}
We choose to set these weight to a simple exponential function of the approximate distances $d_c$ used by the multi-sequence algorithm.  

Similarities are evaluated for every vector in list $\cL_c$, for every cell $c$ returned by the multi-sequence algorithm. In~\cite{BaLe12}, search is terminated if a quota of at least $T$ vectors have been evaluated. For a query vector $\vy$, we may assume that cells $\{c^1, \ldots, c^W \}$ were visited by the multi-index algorithm until termination. Let that sequence of lists visited be $\cL^{\vy} = \{\cL_{c^1}, \dots, \cL_{c^W} \}$. Also let $\cX^{\vy} = \{ \vx_1 ,\dots,\vx_T \}$ be the sequence of the $T$ database vectors evaluated during search, concatenated from $W$ disjoint inverted lists. 
Let also set $\cS^{\vy} = \{ \sigma_w(\vy, \vx_1), \dots, \sigma_w( \vy, \vx_T) \}$, hold the similarity values of the query vector $\vy$ with each of the top $T$ database vectors returned by the multi-index. The ranked list of nearest neighbors returned for the query is the top elements of $\cS^{\vy}$ in descending order according to the approximate similarity values. 




\subsection{Searching with large query sets}
\label{subsec:recommendations}

\begin{algorithm}[t]
\footnotesize{
\DontPrintSemicolon
\SetFuncSty{textsc}
\SetDataSty{emph}
\newcommand{\commentsty}[1]{{\color{green!50!black}#1}}
\SetCommentSty{commentsty}
\SetKwInOut{Input}{input}
\SetKwInOut{Output}{output}

\Input{queries, documents: flattened query and database files. Each row contains some id and a triplet of a fine code, its position in the ordered set and its corresponding coarse code.}
\Output{scores: list of documents sorted by LOH similarity to the query set}
\BlankLine
\tcp{load flattened files for query set}
Q = LOAD 'queries' AS (\texttt{user\_id}, \texttt{image\_id\_q}, (\texttt{coarse\_code,position,fine\_code}) as \texttt{code\_q});

\BlankLine
\tcp{load flattened files for the database set}
D = LOAD 'documents' AS (\texttt{image\_id\_d}, (\texttt{coarse\_code,position,fine\_code}) as \texttt{code\_d});

\BlankLine
\tcp{join by code}
matches = JOIN Q by \texttt{code\_q}, D BY \texttt{code\_d};

\BlankLine
\tcp{group by document}
grouped = GROUP matches BY (\texttt{user\_id}, \texttt{image\_id\_d});

\BlankLine
\tcp{count the matches within each group}
scores = FOREACH grouped GENERATE 
	group.\$0 as \texttt{user\_id},
	group.\$1 as \texttt{image\_id},
	COUNT(matches) as \texttt{n\_matches};

\BlankLine
\tcp{order by number of matches}
scores = ORDER scores BY \texttt{user\_id} ASC, \texttt{n\_matches} DESC;

\caption{Pseudo-code for batch search in PIG}
\label{algo:pig}
}
\end{algorithm}

\noindent
An application like recommendation requires the ability to jointly search with multiple query vectors and get aggregated results. Taking visual recommendation as an example, one can define query sets in multiple ways, \eg the set of images in a given Flickr group, or the set of images that a given user has favorited. If each image is represented as a vector, \eg using CNN-based global visual features, queries with \emph{image sets} correspond to queries with \emph{multiple vectors} and can produce results that are visually similar to the whole query image set.



Now let's suppose that the query is a set of $Y$ vectors, $\cY = \{ \vy_1, \dots, \vy_Y\}$. 
If we query the index for each of the query vectors, we get sets $\cL^{\vy}$, $\cX^{\vy}$ and $S^{\vy}$ for each vector corresponding to cells, database vectors and similarities, respectively, with  ${\vy} = 1, \dots, Y$. We can now define the set of similarity values
\begin{equation}
	S_{\cY} = \{ \sigma(\cY, \vx_t) \} \text{ for } t \in \bigcup_{{\vy} \in Y} \cX^\vy,
\label{eq:mqsim}
\end{equation}
between the query set $\cY$ and all database vectors evaluated in any of the single-vector queries. The aggregation function $\sigma(\cY, \vx) = g( \sigma_w(\vy_1, \vx), \dots, \sigma_w(\vy_Y, \vx) )$ measures the similarity of the query set $\cY$ to database vector $\vx$, where $g$ can be any pooling function, for example
\begin{equation}
	\sigma_{SUM}(\cY, \vx)= \sum_{\vy \in \cY} \sigma_w(\vy, \vx),
\label{eq:mqsum}
\end{equation}
for sum-pooling or 
$\sigma_{MAX}(\cY, \vx)= \arg \max_{\vy \in \cY} \sigma_w(\vy, \vx)$
for max-pooling.  We experimentally found that sum-pooling performs better than max-pooling, which is understandable since the latter tends to under-weight results that appeared in the result sets of many query vectors. We also experimented with more complex functions, \eg functions that combine max-pooling with the frequency that each database image appears in the result sets, but since the improvements were minimal we end up using the simpler sum-pooling function of (\ref{eq:mqsum}) for the rest of the paper.

An advantage of representing images as multiple hash codes is that they can be naturally manipulated for a variety of tasks with MapReduce using tools such as PIG or HIVE for Hadoop.
Returning to our example of image recommendations in Flickr, we might use a user's favorited images as a query set to produce a ranked set of recommended images that are visually similar to images the user has favorited. In this case we would like to produce recommendations for \textit{all} users, and we would like an algorithm that can run many searches efficiently in batch to periodically recompute image recommendations for all users.

We show PIG pseudo-code for such a batch search in Algorithm~\ref{algo:pig}. The algorithm assumes that the coarse and fine codes for each document have already been computed and are available in a \textit{flattened} form. To get this form, we first split the $m$ fine codes and create triplets by appending each fine code with its position in $f$ and the corresponding coarse code. That is, for coarse and fine codes $(c_1, c_2)$ and $(f_1, \ldots, f_m)$, respectively, we would get the set of $m$ triplets $( (c_1, 1, f_1)$, $(c_1, 2, f_2),$ $\ldots, (c_2, m, f_m) )$ or \textit{LOH codes}. 




\subsection{Clustering and deduplication with LOH}
\label{subsec:dedup}

\noindent
LOH can also be used for efficient and scalable clustering. 
Unlike recent approaches that try to cluster data on a single machine~\cite{iccv15, gong15}, we are interested in the distributed case.
Our clustering algorithm first constructs a graph of documents from an input set $\cY$ such that a pair of documents $\vx, \vy \in \cY$ is connected by an edge iff the LOH similarity of the pair is above some threshold $t$, \ie $\sigma_h(\vy, \vx) > t$. Clustering then amounts to finding connected components in this graph.

We present pseudo-code for LOH clustering in Algorithm~\ref{alg:dedup}. The algorithm first groups documents by flattened code triplets. This grouping is used to efficiently count the number of matches for document pairs by greatly reducing the number of pairs we consider when constructing the graph. Like the batch search algorithm, LOH clustering is easily implemented in MapReduce frameworks.
When running with a high threshold for a small set of images, \eg for the top hundred or thousand results after an image search, this algorithm can deduplicate the set in real-time, requiring only a few milliseconds.

\begin{algorithm}[t]
\footnotesize{
\DontPrintSemicolon
\SetFuncSty{textsc}
\SetDataSty{emph}
\newcommand{\commentsty}[1]{{\color{green!50!black}#1}}
\SetCommentSty{commentsty}
\SetKwInOut{Input}{input}
\SetKwInOut{Output}{output}
\Input{documents $\cY = \{\vy_1, \ldots, \vy_n\}$ with the set of flattened triplets, threshold $t$}
\Output{clusters $\cR =\{r_1, \ldots, r_n\}$}
\BlankLine

Groups = HashMap$<$List$>$()\;
Matches = HashMap$<$Int$>$()\;
DocumentGraph = Graph()\;

\BlankLine
\tcp{group documents by matching LOH codes}
\ForEach{$d \in D$}
{
  \ForEach{$(c_i, j, f_j) \in d$}
  {
    Groups($(c_i, j, f_j)$).append($d$)\;
  }
}

\BlankLine
\tcp{count num of matching codes for pairs of documents in each group}
\ForEach{$group \in Groups$}
{
  \ForEach{$(d_a,d_b) \in allPairs(group)$}
  {
    Matches($(d_a,d_b)$)++\;
    \If{Matches($(d_a,d_b)$) $>$ t}
    {
      DocumentGraph.addEdge($d_a,d_b$) \;
    }
  }
}

\BlankLine
\tcp{find connected components in the document graph}
$\cR \longleftarrow$ DocumentGraph.findConnectedComponents() \;
\Return{$\cR$}

\caption{Pseudo-code for LOH clustering}
\label{alg:dedup}
}
\end{algorithm}



\section{Experiments}
\label{sec:experiments}

\noindent

\noindent
We use the following four datasets:

\noindent
\textbf{SIFT1M dataset}~\cite{JeDS11}. This dataset contains of 1 million 128-dimensional SIFT vectors and 10K query vectors and is a common benchmark dataset in related work~\cite{KaAv14, NoPF12, WaKC10}.

\noindent
\textbf{Yahoo Flickr Creative Commons 100M dataset}~\cite{TSF+15}. 
This dataset (YFCC100M) contains a subset of 100 million public images with a creative commons license from Flickr and is the largest such publicly available collection of social multimedia images.

\noindent
\textbf{Flickr Brad Pitt Search Dataset\label{dset:bpitt}}.
This dataset contains the top 1048 photo results from a query for the search term "Brad Pitt" on the Flickr website. At the time that we collected this data, results from this query exhibited a large amount of "near-duplicate" results. We manually grouped each photo that looked visually similar into the same cluster. The dataset contains a total of 30 clusters with more than 1 image.

\noindent
\textbf{7 Flickr Groups dataset}. This dataset contains $70K$ images in total and was constructed by collecting $10K$ images from $7$ popular Flickr groups: \emph{Graffiti of the world}, \emph{Sailboats and sailing}, \emph{Glaciers, Icefields and Icebergs},\emph{Windmills}, \emph{Columns and Columns}, \emph{Vintage Cars and Trucks} and \emph{Portraits and Faces}. 




\begin{figure*}[t]
\begin{subfigure}{.36\linewidth}
\begin{tikzpicture}
\begin{semilogxaxis}[%
	height=4.5cm,
	width=\linewidth,
	xlabel={$R$},
	ylabel={recall@$R$},
	legend pos=south east,
	legend style={font=\fontsize{4}{5}\selectfont},
	minor y tick num=1,
]
	\pgfplotstableread{
		r itq lsh  usplh  loh 
		52	0.245 nan nan nan 
		58	nan 0.14 nan nan	    
		67 	nan nan 0.33 nan
		63 nan nan nan 0.28
		113 nan nan nan 0.40
		130 0.360 0.20 nan nan
		132 nan nan nan 0.436
		140 nan nan 0.43 nan
		213 nan nan nan 0.537
		260 nan 0.28 nan nan 
		270 0.50 nan nan nan 
		272 nan nan nan 0.591
		280 nan nan 0.53 nan
		473 nan nan nan 0.705
		500 nan 0.36 nan nan
		540 0.63 nan nan nan
		590 nan nan 0.64 nan
		670 nan nan nan 0.772
		901 nan nan nan 0.817
		910 0.81 nan 0.795 nan
		950 nan 0.46 nan nan
		1000 nan nan nan 0.834
	}{\map}
	\addplot[black,mmark]   table[y=lsh]  \map; \leg{LSH~\cite{DIIM04}}
	\addplot[green!70!black,mmark]   table[y=itq]     \map; \leg{ITQ~\cite{GoLa11}}
	\addplot[magenta,mmark]     table[y=usplh]     \map; \leg{USPLH~\cite{WaKC10}}
	\addplot[cyan,mmark]    table[y=loh]    \map; \leg{LOH (ours)}
\end{semilogxaxis}
\end{tikzpicture}
\caption{}
\label{fig:hashsift1m}
\end{subfigure} 
\hspace{-12pt}
\begin{subfigure}{.32\linewidth}
\begin{tikzpicture}
\begin{semilogxaxis}[%
	width=\linewidth,
	height=4.5cm,
	xlabel={$R$},
	ylabel={recall@$R$},
	legend pos=south east,
	legend style={font=\fontsize{4}{5}\selectfont},
	minor y tick num=1,
]
	\pgfplotstableread{
		r lopq loh sh
		10     0.85 0.23 0.23
		100    0.98 0.49 0.48
		1000   0.98 0.86 0.82
		10000  0.99 0.99 0.95
	}{\map}
	\addplot[green!70!black,mmark]   table[y=sh]     \map; \leg{SH~\cite{WeTF08}}
	\addplot[cyan,mmark]    table[y=loh]    \map; \leg{LOH (ours)}
\end{semilogxaxis}
\end{tikzpicture}
\caption{}
\label{fig:sift1m}
\end{subfigure}
\hspace{-12pt}
\begin{subfigure}{.36\linewidth}
\begin{tikzpicture}
\begin{axis}[%
	height=4.5cm,
	width=\linewidth,
	xlabel={$P$},
	ymin=0.3,
	ylabel={precision@$P$},
	legend pos=south east,
	legend style={font=\fontsize{4}{5}\selectfont},
	minor y tick num=1,
]
	\pgfplotstableread{
		r col gla gra sail vin wind  
		10  0.9 1 1 1 0.9 0.7
		100 0.89 0.96 0.97 0.99 0.98 0.79
		200 0.845 0.965 0.97 0.985 0.985 0.685
		300 0.86333333333333 0.9666 0.97333 0.99 0.99 0.6
		400 0.89 0.9675 0.9725 0.99 0.9875 0.505
		500 0.904 0.962 0.976 0.988 0.99 0.492
	}{\map}
	\addplot[green!70!black,mmark]   table[y=col]     \map; \leg{columns}
	\addplot[red,mmark]     table[y=gla]     \map; \leg{glaciers}
	\addplot[blue,mmark]    table[y=gra]    \map; \leg{graffiti}
	\addplot[yellow!60!black,mmark]    table[y=sail]    \map; \leg{sailing}
	\addplot[magenta,mmark]    table[y=vin]    \map; \leg{vintage cars}
	\addplot[cyan,mmark]    table[y=wind]    \map; \leg{windmills}
\end{axis}
\end{tikzpicture}
\caption{}
\label{fig:groupstats}
\end{subfigure}

\caption{\textit{Left}: Recall@$R$ on SIFT1M with 64bit codes. Recall is measured here as the percentage of times the true nearest neighbor is returned within the top $R$ results returned by the index for all $10K$ queries. $K=1024$ for LOH. 
\textit{Center}: Recall@$R$ on SIFT1M for the proposed LOH and Spectral Hashing. LOH only takes into account the top $T=10000$ results returned by the multi-index, while SH is exhaustive.
\textit{Right}: Precision@$P$ for the 7 Flickr Group Photos dataset. Recommendations for the group ``Portraits \& Faces'' are not depicted because they were flawless.}
\label{fig:somename}
\end{figure*}
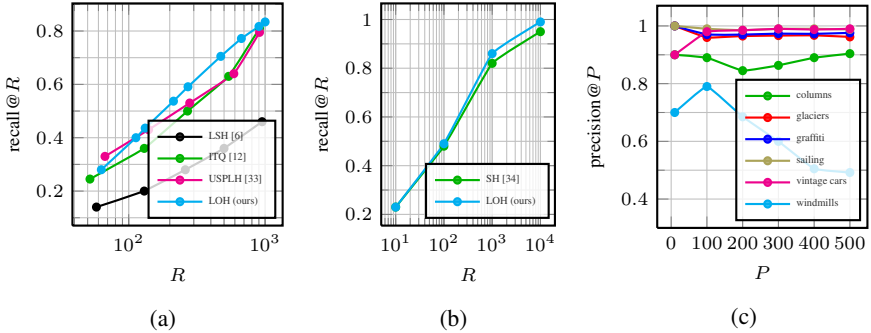


We use the $fc7$ features of the pretrained AlexNet model~\cite{KrSH12} from Caffe~\cite{YSD+14} and use PCA to reduce them to 128 dimensions. We learn a covariance matrix from $100$ million images of the YFCC100M dataset and further permute the dimensions in order to balance variance between the two subspaces before multi-indexing~\cite{GHKS14}. 
As in all related work~\cite{JeDS11, GHKS14, KaAv14}, we set the number of sub-quantizer centroids $k = 256$, \ie we require $m$ bytes of memory in total per vector.

We adopt the Multi-LOPQ~\cite{KaAv14} approach to train\footnote{\url{https://github.com/yahoo/lopq}} and index the database points. We use parameters $K=1024$ ($K=8192$), $m=8$ ($m=16$) and for $T=10K$ ($T=100K$) for SIFT1M (YFCC100M).

We use the recall metric to measure the performance of LOH against related methods and conduct experiments on the SIFT1M dataset. To compare with hashing methods that do not return any ordering of the retrieved points, we measure the percentage of times the true nearest neighbor is within the top $R$ results returned by the index (therefore varying parameter $T$ of the multi-index for LOH) for all $10K$ queries of the SIFT1M dataset. We use the precision-recall metric for evaluating deduplication.


\subsection{Approximate nearest neighbor search with LOH}
\label{subsec:hash}


\noindent
We first investigate how the LOH approach compares with the hashing literature for the task of retrieving the true nearest neighbor within the first $R$ samples seen. We compare against classic hashing methods like Locality Sensitive Hashing (LSH) \cite{DIIM04}, Iterative Quantization \cite{GoLa11} and the recent Sequential Projection Learning Hashing (USPLH)~\cite{WaKC10} and report results in Figure \ref{fig:hashsift1m}. One can see that LOH, built on the inverted multi-index after balancing the variance of the two subspaces, compares well with the state-of-the-art in the field, even outperforming recently proposed approaches like \cite{WaKC10} for large enough $R$. 

In~\ref{fig:sift1m}, we evaluate LOH ranking, \ie how well LOH orders the true nearest neighbor after looking at a fraction of the database. We compare against Spectral Hashing (SH)~\cite{WeTF08} which, like LOH, also provides a ranking of the results. LOH performs similarly for small values of $R$ but outperforms SH when retrieving more than $R=100$ results, which is the most common case.




\subsection{Visual recommendations for Flickr groups}
\label{subsec:group}

We conduct an experiment to evaluate the ability of the proposed approach to visually find images that might be topically relevant to a group of photos already curated by a group of users. On Flickr, such activity is common as users form groups around topical photographic interests and seek out high-quality photos relevant to the group. Group moderators may contact photo owners to ask them to submit to their group.

To evaluate this, we select 7 public Flickr groups that are representative of the types of topical interests common in Flickr groups, selected due to their clear thematic construction (\emph{graffiti}, \emph{sailing}, \emph{glacier}, \emph{windmill}, \emph{columns}, \emph{cars \& trucks}, \emph{portrait \& face}), for ease of objective evaluation. For each group, we construct a large query of $10,000$ images randomly sampled from the group pool.
We perform visual search using our proposed method on the YFCC100M dataset, aggregate results from all 10 thousand images and report precision after manual inspection of the top $k=500$ results. We visually scanned the photo pools of the groups and consider true positives all images that look like images in the photo pool and follow the group rules as specified by the administrators of each group.

\begin{figure*}[t!]
	\centering
    \begin{subfigure}[t]{.49\linewidth}
		\centering\includegraphics[width=\linewidth]{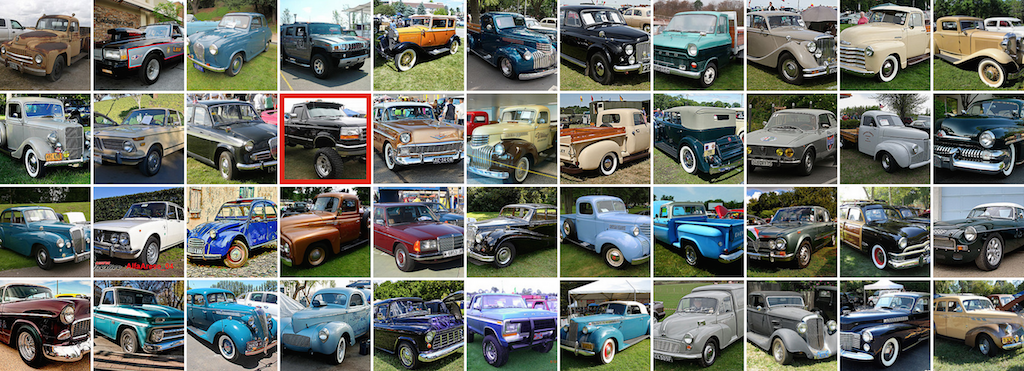}
	\caption{LOH-based results (proposed)}
	\label{fig:vintageours}
	\end{subfigure}
    \begin{subfigure}[t]{.49\linewidth}
		\centering\includegraphics[width=\linewidth]{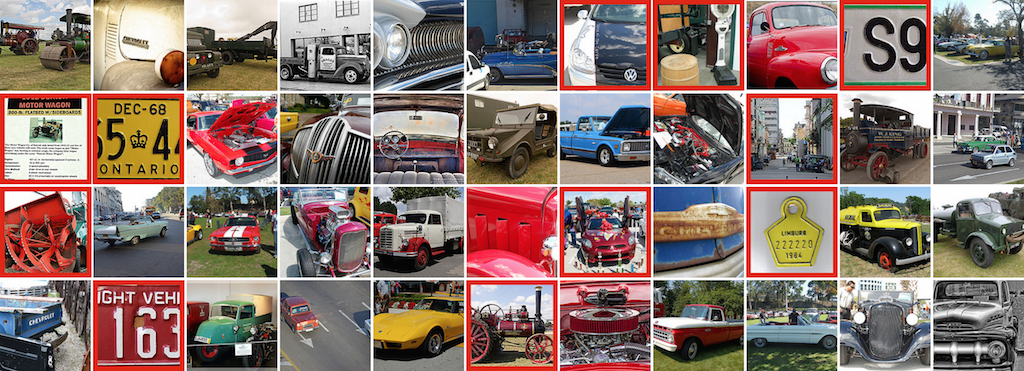}
	\caption{Tag-based results (baseline)}
	\label{fig:vintageflickr}
	\end{subfigure}

	\caption{Suggestions for Flickr group ``vintage cars and trucks'', false positives are marked by a red border. Figure~\ref{fig:vintageours}: Our visual similarity-based suggestions; only 1 of the top results is false and the aesthetics fit the images in the group. Figure~\ref{fig:vintageflickr}: Top images returned by tag-based search for ``vintage cars and trucks'';  we see more false positives in this case.}
	\label{fig:vintage}
\end{figure*}

Precision for each group is shown in Figure~\ref{fig:groupstats}. For group ``Portraits \& Faces'', for example \emph{all} 500 top results were high-quality portraits. We see some confusion due to the nature of the visual representation chosen (\eg our visual representation may confuse desert and cloud images with snow images), but overall, Precision@500 was over $0.96$ for five out of the seven groups we tested. 





Example results for the set expansion with our method and a baseline tag-based search are shown in Figure \ref{fig:vintage} for Flickr group ``vintage cars and trucks''. For the proposed approach, precision is high, as is the aesthetic quality of the results. The tag-based Flickr search returns more false positives for such a specific group, as irrelevant images are likely improperly tagged.

\subsection{Clustering and deduplication results}
\label{subsec:dedupexp}

We evaluate the performance of LOH on the deduplication task using a dataset of Flickr searches for the query ``Brad Pitt'' with the LOH codes learned on the YFCC100M dataset. To measure precision and recall, we enumerate all pairs of images in the dataset and define a ``positive'' sample as a pair of images that belong to the same group in our dataset, and a ``negative'' sample as a pair of images that belong to different groups in our dataset.

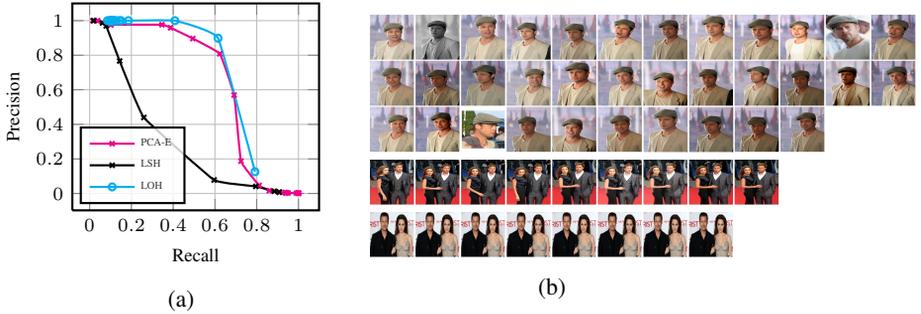
\begin{figure*}[t]

\begin{subfigure}[c]{.4\linewidth}
\begin{tikzpicture}
	\begin{axis}[
		height=4.3cm,
		xlabel=Recall,
		ylabel=Precision,
        legend pos=south west,
		legend style={font=\fontsize{4}{5}\selectfont}
    ]
	
	\addplot[color=magenta,mark=x] coordinates {
		(0.015702, 1.000000)
		(0.038843, 1.000000)
		(0.101653, 0.976190)
		(0.101653, 0.976190)
		(0.346281, 0.976690)
		(0.387603, 0.959100)
		(0.494215, 0.896552)
		(0.623140, 0.808146)
		(0.691736, 0.569388)
		(0.724793, 0.187193)
		(0.811570, 0.046934)
		(0.860331, 0.016420)
		(0.906612, 0.007390)
		(0.935537, 0.005394)
		(0.940496, 0.003965)
		(0.940496, 0.003965)
		(0.949587, 0.003018)
		(0.990083, 0.002337)
		(0.999174, 0.002251)
		(1.000000, 0.002218)
		(1.000000, 0.002210)
		(1.000000, 0.002206)
	};
	\addlegendentry{PCA-E}
    
	\addplot[color=black,mark=x] coordinates {
		(0.0190,1.0000)
		(0.0612,0.9867)
		(0.0793,0.9697)
		(0.1438,0.7665)
		(0.2595,0.4398)
		(0.5959,0.0779)
		(0.7959,0.0411)
		(0.8843,0.0134)
		(0.9083,0.0080)
	};
	\addlegendentry{LSH}
    
	\addplot[color=cyan,mark=o] coordinates {
		(0.791735537184, 0.125820856317)
		(0.614876033053, 0.898550724627)
		(0.408264462807, 0.99999999998)
		(0.185950413222, 0.999999999956)
		(0.149586776858, 0.999999999945)
		(0.142148760329, 0.999999999942)
		(0.127272727272, 0.999999999935)
		(0.118181818181, 0.99999999993)
		(0.113223140495, 0.999999999927)
		(0.111570247933, 0.999999999926)
		(0.105785123966, 0.999999999922)
		(0.105785123966, 0.999999999922)
		(0.102479338842, 0.999999999919)
		(0.0983471074372, 0.999999999916)
		(0.0958677685942, 0.999999999914)
		(0.0851239669414, 0.999999999903)
	};
	\addlegendentry{LOH}


	\end{axis}
\end{tikzpicture}
\caption{}
\label{fig:prbpitt}
\end{subfigure}
\begin{subfigure}[c]{.4\linewidth}
\centering
\includegraphics[height=3.2cm]{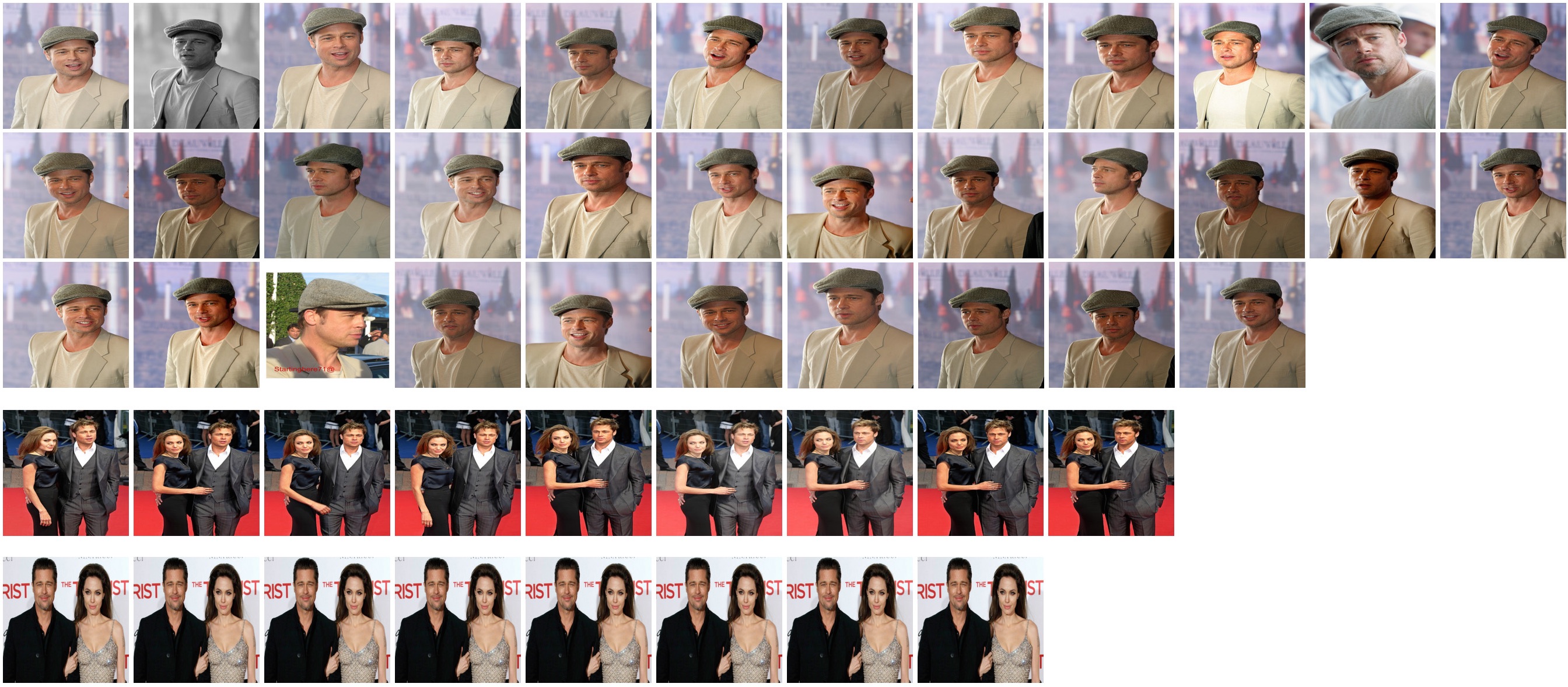}
\caption{}
\label{fig:brad}
\end{subfigure}

\caption{\textit{Left:} Precision-Recall curve on the Brad Pitt dataset. \textit{Right:} Deduplication results on the Brad Pitt dataset; three clusters of duplicates are shown where each cluster shares at least $3$ LOH codes.}
\end{figure*}

In figure~\ref{fig:prbpitt} we plot the precision-recall for LOH versus LSH~\cite{DIIM04} and PCA-E~\cite{GPGL+2011}. For LSH, we transform our PCA'd 128 dimensional image descriptor into a 128-bit binary code computed from random binary projection hash functions. For PCA-E we compute a 128-bit binary code by subtracting the mean of our PCA'd 128 dimensional image descriptor and taking the sign.


\noindent

\begin{figure*}[t]
	\centering
	\includegraphics[width=.31\linewidth]{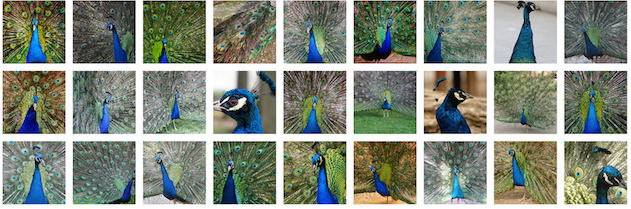}
    \hspace{6pt}
	\includegraphics[width=.31\linewidth]{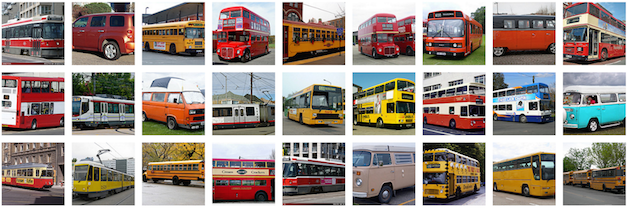}
    \hspace{6pt}
	\includegraphics[width=.31\linewidth]{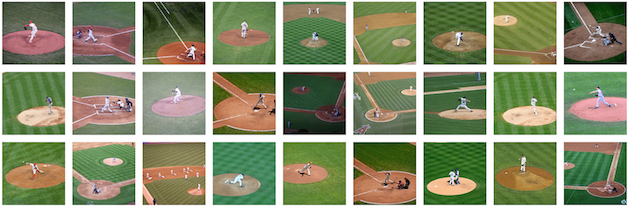}

\vspace{8pt}
	\includegraphics[width=.31\linewidth]{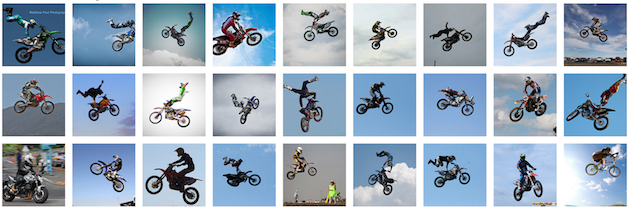}
    \hspace{6pt}
	\includegraphics[width=.31\linewidth]{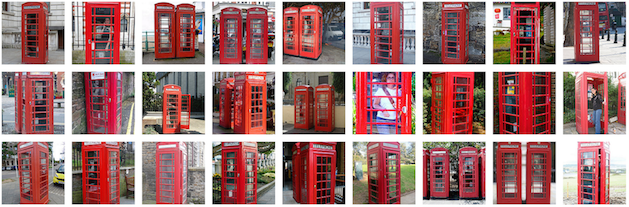}
    \hspace{6pt}
	\includegraphics[width=.31\linewidth]{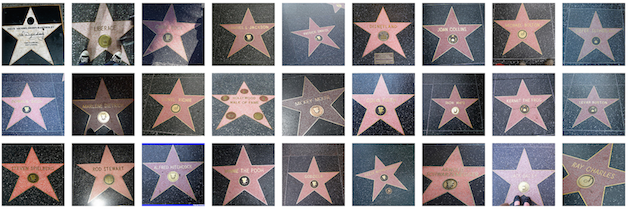}

\vspace{8pt}
	\includegraphics[width=.31\linewidth]{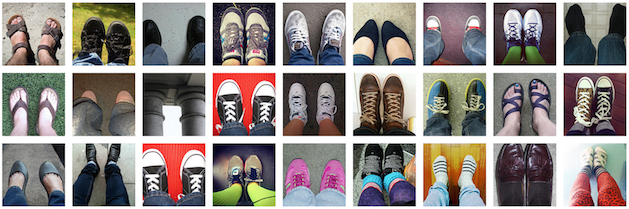}
    \hspace{6pt}
	\includegraphics[width=.31\linewidth]{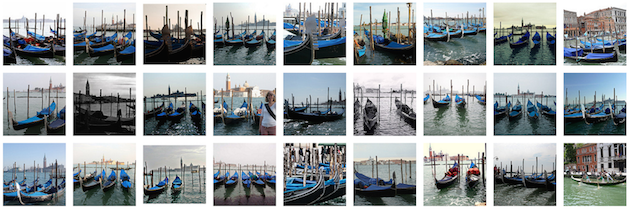}
    \hspace{6pt}
	\includegraphics[width=.31\linewidth]{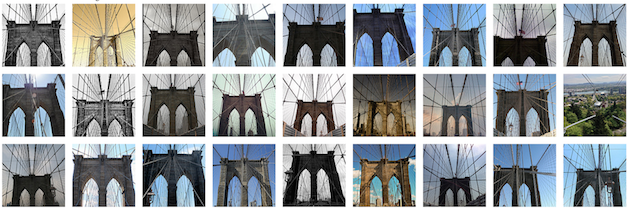}

\vspace{13pt}
	\includegraphics[width=.31\linewidth]{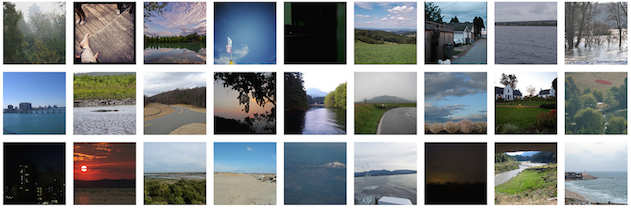}
    \hspace{6pt}
	\includegraphics[width=.31\linewidth]{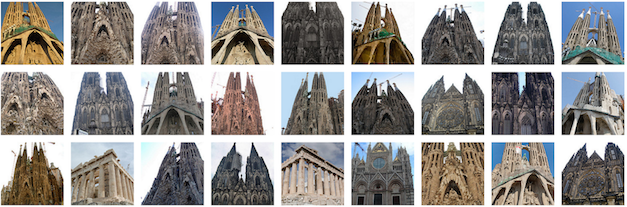}
    \hspace{6pt}
	\includegraphics[width=.31\linewidth]{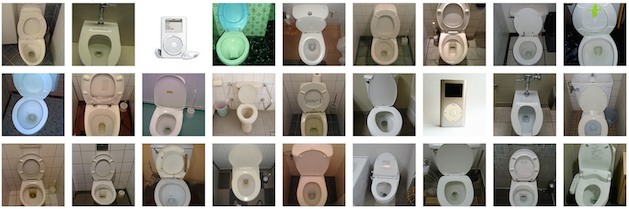}
	\caption{Random images from sample clusters of the YFCC100M dataset. The first three rows show (mostly) coherent clusters that can be used for learning classifiers, while the bottom row shows failure cases. The sizes of the clusters depicted are (row-wise): 1194,976,920 / 272,320,873 / 139,151,164 / 6.8M, 1363 and 94.}
	\label{fig:yfccclusters}
\end{figure*}

We run LOH clustering for the 100 million images of the YFCC100M dataset on a small Hadoop cluster and show sample clusters in Figure~\ref{fig:yfccclusters}. We first did some cleaning and preprocessing by using a stoplist of codes (\ie remove all triplets that appear more than $10K$ or less than $10$ times) for efficiency. For a threshold of $t=3$, we get $74$ million edges and approximately $7$ million connected components. Of those, about 6.5 million are small components of size smaller than $5$. 
The graph construction for the 100M images took a couple of hours, while connected components runs in a few minutes.

By visual inspection, we notice that a large set of medium-sized clusters (\ie clusters with $10^2 - 10^4$ images) contain visually consistent higher level concepts (\eg from Figure~\ref{fig:yfccclusters}: ``motorbikes in the air'', ``Hollywood St stars'' or  ``British telephone booths''). Such clusters can be used to learn classifiers in a semi-supervised framework that incorporates noisy labels. Clustering YFCC100M gives us about 32K such clusters.


\section{Conclusions}
\label{sec:discussion}

\noindent
In this paper we propose a novel matching scheme, Locally Optimized Hashing or LOH, that is computed on the very powerful and compact LOPQ codes. We show how LOH can be used to efficiently perform visual search, recommendation, clustering and deduplication for web-scale image databases in a distributed fashion.

While LOPQ distance computation gives high quality, fast distance estimation for nearest neighbor search, it is not as well suited for large-scale, batch search and clustering tasks. LOH, however, enables these use-cases by allowing implementations that use only highly parallelizable set operations and summations. LOH can therefore be used for massively parallel visual recommendation and clustering in generic distributed environments with only a few lines of code. Its speed also allows its use for deduplication of, \eg, search result sets at query time, requiring only a few milliseconds to run for sets of thousands of results.




\bibliography{tex/bib}

\begin{thebibliography}{10}
\providecommand{\url}[1]{\texttt{#1}}
\providecommand{\urlprefix}{URL }

\bibitem{ArZi12b}
Arandjelovic, R., Zisserman, A.: Multiple queries for large scale specific
  object retrieval. In: BMVC (2012)

\bibitem{iccv15}
Avrithis, Y., Kalantidis, Y., Anagnostopoulos, E., Emiris, I.Z.: Web-scale
  image clustering revisited. In: ICCV (2015)

\bibitem{AKTS10}
Avrithis, Y., Kalantidis, Y., Tolias, G., Spyrou, E.: Retrieving landmark and
  non-landmark images from community photo collections. In: ACM Multimedia
  (2010)

\bibitem{BaLe12}
Babenko, A., Lempitsky, V.: The inverted multi-index. In: CVPR (2012)

\bibitem{BSCL14}
Babenko, A., Slesarev, A., Chigorin, A., Lempitsky, V.: Neural codes for image
  retrieval. In: ECCV (2014)

\bibitem{DIIM04}
Datar, M., Immorlica, N., Indyk, P., Mirrokni, V.: Locality-sensitive hashing
  scheme based on p-stable distributions. In: Symposium on Computational
  Geometry (2004)

\bibitem{DABP14}
Doll{\'a}r, P., Appel, R., Belongie, S., Perona, P.: Fast feature pyramids for
  object detection. PAMI  36(8),  1532--1545 (2014)

\bibitem{DoJP16}
Douze, M., J{\'e}gou, H., Perronnin, F.: Polysemous codes. In: ECCV (2016)

\bibitem{FeTu13}
Fernando, B., Tuytelaars, T.: Mining multiple queries for image retrieval:
  On-the-fly learning of an object-specific mid-level representation. In: ICCV
  (2013)

\bibitem{GHKS14}
Ge, T., He, K., Ke, Q., Sun, J.: Optimized product quantization. Tech. Rep.~4
  (2014)

\bibitem{GDDM14}
Girshick, R., Donahue, J., Darrell, T., Malik, J.: Rich feature hierarchies for
  accurate object detection and semantic segmentation. In: CVPR (2014)

\bibitem{GoLa11}
Gong, Y., Lazebnik, S.: Iterative quantization: A procrustean approach to
  learning binary codes. In: CVPR (2011)

\bibitem{gong15}
Gong, Y., Pawlowski, M., Yang, F., Brandy, L., Bourdev, L., Fergus, R.: Web
  scale photo hash clustering on a single machine. In: CVPR (2015)

\bibitem{GPGL+2011}
Gordo, A., Perronnin, F., Gong, Y., Lazebnik, S.: Asymmetric distances for
  binary embeddings. PAMI  36(1),  33--47 (2014)

\bibitem{HsCA14}
Hsiao, K., Calder, J., Hero, A.O.: Pareto-depth for multiple-query image
  retrieval. arXiv preprint arXiv:1402.5176  (2014)

\bibitem{JeDS11}
J\'egou, H., Douze, M., Schmid, C.: Product quantization for nearest neighbor
  search. PAMI  33(1) (2011)

\bibitem{JTDA11}
J\'egou, H., Tavenard, R., Douze, M., Amsaleg, L.: Searching in one billion
  vectors: Re-rank with source coding. In: ICASSP (2011)

\bibitem{YSD+14}
Jia, Y., Shelhamer, E., Donahue, J., Karayev, S., Long, J., Girshick, R.,
  Guadarrama, S., Darrell, T.: Caffe: Convolutional architecture for fast
  feature embedding. arXiv preprint arXiv:1408.5093  (2014)

\bibitem{JHL+13}
Jin, Z., Hu, Y., Lin, Y., Zhang, D., Lin, S., Cai, D., Li, X.: Complementary
  projection hashing. In: ICCV (2013)

\bibitem{KTA+11}
Kalantidis, Y., Tolias, G., Avrithis, Y., Phinikettos, M., Spyrou, E., Mylonas,
  P., Kollias, S.: Viral: Visual image retrieval and localization. MTAP  (2011)

\bibitem{KaAv14}
Kalantidis, Y., Avrithis, Y.: Locally optimized product quantization for
  approximate nearest neighbor search. In: CVPR (2014)

\bibitem{KNA+07}
Kennedy, L., Naaman, M., Ahern, S., Nair, R., Rattenbury, T.: How flickr helps
  us make sense of the world: Context and content in community-contributed
  media collections. In: ACM Multimedia. vol.~3, pp. 631--640 (2007)

\bibitem{KrSH12}
Krizhevsky, A., Sutskever, I., Hinton, G.E.: Imagenet classification with deep
  convolutional neural networks. In: NIPS (2012)

\bibitem{NoFl13}
Norouzi, M., Fleet, D.: Cartesian $k$-means. In: CVPR (2013)

\bibitem{NoPF12}
Norouzi, M., Punjani, A., Fleet, D.J.: Fast search in {Hamming} space with
  multi-index hashing. In: CVPR (2012)

\bibitem{PaJA10}
Paulev{\'e}, L., J{\'e}gou, H., Amsaleg, L.: {Locality sensitive hashing: a
  comparison of hash function types and querying mechanisms}. Pattern
  Recognition Letters  31(11),  1348--1358 (Aug 2010)

\bibitem{RDS+14}
Russakovsky, O., Deng, J., Su, H., Krause, J., Satheesh, S., Ma, S., Huang, Z.,
  Karpathy, A., Khosla, A., Bernstein, M., et~al.: Imagenet large scale visual
  recognition challenge. arXiv preprint arXiv:1409.0575  (2014)

\bibitem{SSLS15}
Shen, F., Shen, C., Liu, W., Shen, H.T.: Supervised discrete hashing. CVPR
  (2015)

\bibitem{TES+16}
Thomee, B., Elizalde, B., Shamma, D.A., Ni, K., Friedland, G., Poland, D.,
  Borth, D., Li, L.J.: Yfcc100m: The new data in multimedia research.
  Communications of the ACM  59(2),  64--73 (2016)

\bibitem{TSF+15}
Thomee, B., Shamma, D.A., Friedland, G., Elizalde, B., Ni, K., Poland, D.,
  Borth, D., Li, L.J.: The new data and new challenges in multimedia research.
  arXiv preprint arXiv:1503.01817  (2015)

\bibitem{ToJA15}
Tolias, G., Avrithis, Y., J{\'e}gou, H.: Image search with selective match
  kernels: Aggregation across single and multiple images. International Journal
  of Computer Vision pp. 1--15 (2015)

\bibitem{WSYY14}
Wang, J., Shen, H.T., Yan, S., Yu, N., Li, S., Wang, J.: Optimized distances
  for binary code ranking. In: Proceedings of the ACM International Conference
  on Multimedia. pp. 517--526. ACM (2014)

\bibitem{WaKC10}
Wang, J., Kumar, S., Chang, S.F.: Sequential projection learning for hashing
  with compact codes. In: ICML (2010)

\bibitem{WeTF08}
Weiss, Y., Torralba, A., Fergus, R.: Spectral hashing. In: NIPS (2008)

\bibitem{XBC+11}
Xu, B., Bu, J., Chen, C., Cai, D., He, X., Liu, W., Luo, J.: Efficient manifold
  ranking for image retrieval. In: SIGIR (2011)

\bibitem{ZZT+13}
Zhang, L., Zhang, Y., Tang, J., Lu, K., Tian, Q.: Binary code ranking with
  weighted hamming distance. In: CVPR (2013)

\bibitem{ZZS+09}
Zheng, Y., Zhao, M., Song, Y., Adam, H., Buddemeier, U., Bissacco, A., Brucher,
  F., Chua, T.S., Neven, H.: Tour the world: Building a web-scale landmark
  recognition engine. In: CVPR (2009)

\bibitem{ZhHS14}
Zhu, C.Z., Huang, Y.H., Satoh, S.: Multi-image aggregation for better visual
  object retrieval. In: ICASSP (2014)

\end{thebibliography}
\bibliographystyle{splncs03}

\end{document}